# AN INSPECTION TECHNOLOGY OF INNER SURFACE OF THE FINE HOLE BASED ON MACHINE VISION

Rongfang He, Weibin Zhang, Guofang Gao*

Institute of Chemical Materials, China Academy of Engineering Physics, Mianyang, Sichuan 621999, China

`herongfang@caep.cn ,gaogf@caep.cn (G. G).`

**ABSTRACT**

*Fine holes are an important structural component of industrial components, and their inner surface quality is closely related to their function.In order to detect the quality of the inner surface of the fine hole,a special optical measurement system was investigated in this paper. A sight pipe is employed to guide the external illumination light into the fine hole and output the relevant images simultaneously. A flexible light array is introduced to suit the narrow space, and the effective field of view is analyzed. Besides, the arc surface projection error and manufacturing assembly error of the device are analyzed, then compensated or ignored if small enough. In the test of prefabricated circular defects with the diameter φ0.1mm, φ0.2mm, 0.4mm distance distribution and the fissure defects with the width 0.3mm, the maximum measurement error standard deviation are all about 10μm. The minimum diameter of the measured fine hole is 4mm and the depth can reach 47mm.*

## 1. INTRODUCTION

With the rapid development of industrialization and advanced manufacturing, the mechanical structure of industrial equipment has become more and more complex. Micro-pipes or small-sized holes have been widely used in energy and chemical engineering, automobile manufacturing, aerospace, medicine, military [1-5]. However, due to the errors in the manufacturing process or the harsh working environment, some pits, bumps, holes, cracks, scratches and other defects often appear on the inner surface of the holes. This may cause leakage of liquid or gas, causing harm to equipment, polluting the environment, and even leading to accidents. It has become an urgent problem in modern industrial manufacturing and quality control to make an online automatic defects measurement of small-sized feature inner holes, pits, cracks, and scratches in micro-pipelines or fine holes accurately and swiftly[6,7].

Many scholars have carried out relevant research on defect detection of the inner surface of micro-holes. At present, two main methods are being widely used, including structured light-based optical inspection methods and image processing-based visual inspection methods.

The structured light detection schemes project laser to the inner surface of small bore and receive signal with camera or light detector. It can be divided into the single-spot laser scanning method and ring-laser section method. Single-spot laser scanning methods usually rely on laser spot and PSD to achieve single point position measurement. Wu, et al. [8] designed a profile sensor based on single-spot laser and realized the reconstruction of long curved pipe. It can be used for pipes with diameter between 9.5mm and 10.5mm, a curvature radius larger than 100mm, and the reconstruction precision reached $\pm0.1$mm. Dong, et al. [9] designed a device set on the CNC repair machine tool to detect the drill pipe thread. In their research, the measuring point inclination angle is analyzed and compensated. The device can measure 25mm thread with an accuracy of $\pm 8\mu m$. Li, et al. [10] designed a distributed layout of the inner diameter detection sensor for automobile engine cylinders, which can measure the shaft holes with diameter of 92mm, and the measurement error is under $4\mu m$. Kulkarni, et al. [11] proposed a new method for measuring the protrusions and depressions on the inner surface of small holes based on laser scattering. This method placed detectors in the reflection and glancing directions respectively. The difference of the signals on the detectors is used to identify defects. This device can obtain an accuracy of 50μm. Ring-laser section methods commonly use the ring-laser and CCD. Wang, et al. [12] proposed a ring discrete distributing laser-spot method to reconstruct the contour of pipe. Ring distributed optical fiber is employed to project the light point, which combined the advantages of single-spot laser scanning and ring-laser section method. Zhang, et al. [13] used structured light projector and CCD to extract the light contours, achieving three-dimensional reconstruction of the inner surface of the small hole. In their schemes, CCD sensors and laser projector placed at each end of the hole, this layout makes it only available for through-hole inspection. Structured light detection methods mainly focus on the reconstruction of contour. For flat defects, its detection ability is limited.

Vision detection solutions are usually implemented by CCD or CMOS image sensors and industrial endoscopes. Fischer, et al. [14] designed a foldable robot with a magnetic wheel to detect the inside of the impeller. Robots equipped with small image sensors can pass through holes no more than 15mm and the inner surface of the measured pipe can be monitored remotely. Hong, et al. [15] used sight-pipe and machine vision technology to measure bores with diameter of 8mm to 15mm, which can identify and extract defects larger than 0.5mm. Wu, et al. [16-18] employed sight-pipe to realize the unification of illumination and image transmission. With the cooperation of motion mechanism, they completed a flexible online measurement of the inner surface of a small hole with a diameter of 10mm. To overcome the requirement of ultra-short-focus lenses in thin lens measurement system, Zavyalov, et al. [19] designed a special hole inspection lens. Through the calculation of lens parameters and analysis of imaging property, an undistorted method is proposed. After correcting the distortion of the image, it can be used to inspect holes between 8mm to 20mm. Gong, et al. [20] proposed a 3D optical measurement system based on point cloud reconstruction and realized the measurement of the inner tooth pitch of the M8 thread and the inner diameter of the large and small threads. However, it would take several hours to collect and analyze data, limited the application in online measurement. Yang, et al. [21] designed an integrated rigid endoscope inspection system and the diameter of the neck reached 6mm. With the help of wave-font coding, the inner surface of workpiece with a diameter of 7mm can be inspected clearly. However, manual operations are needed and quantitative results are not provided.

The researches above rely on complicated mechanical mechanisms or can only detect through holes. Moreover, the above-mentioned structure can only detect holes over 6mm, and the requirements for measurement conditions are relatively high. Besides, most of these methods have limited ability to recognize fine features and are mainly used for qualitative measurement. Therefore, we designed a measurement device that imports the external illumination into the hole and export the images of the inner surface. It can realize the fine feature measurement of counterbore or through-hole with diameter between 4mm to 6mm, and depth no more than 47mm. The layout of the remaining parts is as follows. Firstly, the structure and operating principle of the system are introduced. Then the arc surface projection error and pixel equivalent calibration errors are compensated respectively. Additionally, the errors introduced by manufacturing and assembly are analyzed. After that, the accuracy of the device is analyzed through the experiments' results of benchmarks. Finally, the main points and the follow-up improvement direction are summarized.

## 2. REAL-TIME MEASUREMENT DEVICE

### 2.1. Structure of the measurement device

The inner surface defects detection system is composed of three parts, imaging part, control part and motion part. The inner surface of hole is illuminated by an LED light source, the reflected light goes through the sight-pipe and microlen to camera. The captured images are processed by feature analysis algorithm. The rotary stage and moving stage give the ending of sight-pipe a motion space with two degrees of freedom, allowing it to move along the hole axis and rotating around it. With the cooperation of control part and motion part, the inner surface of the measured hole is separated into several parts. After image processing, the measured features would be generated.

The principle of external illumination importation and internal image export to imaging system is shown as Fig. 1. We employ sight-pipe as the light guiding medium. The diameter of the neck is only 3.08mm and the length reaches 52mm. The head of the sight-pipe is composed of two inverted round tables for easy installation, positioning, and the introduction of external light sources. The slender and regular neck makes it easy to enter the thin tube. A 45° reflecting mirror is mounted at the end, and it can be used to deflect the optical path. The illuminating light entered the surface of sight-pipe's head, transmitted through the thin neck, turned at the mirror, and lightened the measured surface. Diffusely reflected beam will be produced by the topography of the measured surface, and again reflected by the mirror, passing through the thin neck and projecting onto the top of the sight-pipe. To reduce the influence of stray light, a shading sleeve is introduced to limit the size of its clear aperture. In addition, we design a flexible LED lighting array to obtain an even illumination.

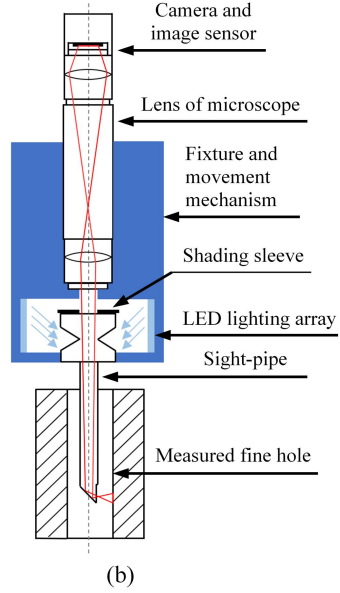

(b)

**Fig. 1** The principles of illuminating and image acquisition.

## 2.2. Measurement path scheme

The imaging system can be modeled by the optical microscope model [22], as shown in Fig. 2. Due to the 45° reflecting mirror, an object on the inner surface of fine hole can be equivalent to its mirror image. The reflecting plane of the sight-pipe can be defined as field diaphragm of the optical system.

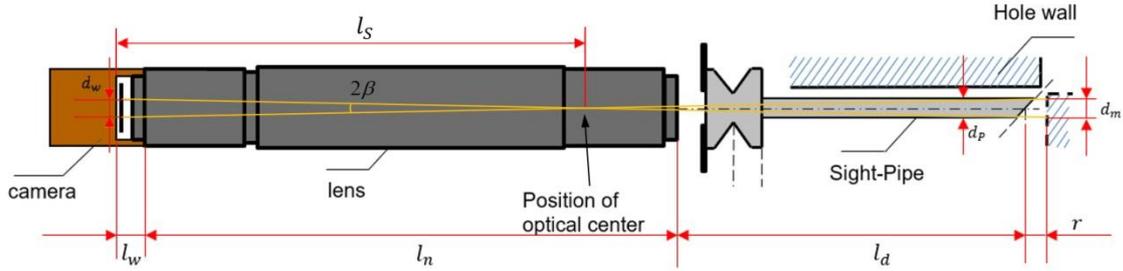

**Fig. 2** Imaging model of the system

Limited by the size of the filed diaphragm, the effective imaging range on the image plane is only a small part, denoted as an area with diameter $d_w$. Assuming the angle formed by the upper and lower sides of this area with the optical center is $2\beta$, and the distance between image plane and optical center is $l_s$, then

$$tan\,\beta = d_w/2\,l_s = d_p/2(l_w+l_n+l_d-l_s) = (d_w+d_p)/2(l_w+l_n+l_d) \quad (1)$$

Where $d_p$ represents the diameter of the reflecting plane, $l_w, l_n, l_d$ represent the distance between image plane and eyepiece, the length of lens, the distance between objective lens and reflecting plane respectively. The diameter of the corresponding object $d_m$ can be calculated as

$$d_m = d_p + 2r\,tan\,\beta = d_p + \frac{r(d_w+d_p)}{l_w+l_n+l_d} \quad (2)$$

In our system, the reflecting plane is a 2.5mm×3.0mm rectangle, the diameter of the measured hole is between 4.0mm and 6.0mm. The distance between image plane and eyepiece $l_w = 15$mm, the length of lens $l_n = 230$mm, the working distance of the lens is $l_d + r$, which ranges from 96mm to 100mm. Typically, the size of the effective image on the image plane is 2.0mm × 2.4mm, and the corresponding size of the hole projected onto the mirror plane is ranging from 2.53mm × 3.03mm to 2.55mm × 3.06mm, depending on the diameter of the measured hole. Due to the bright and dark jumps of the sight-pipe boundary and imaging distortion (described in Sec.3.1), we only take the area of 1.5mm × 1.5mm in the center of the image plane in the experiment, noted as $d_r \times d_h$.].

Now that a single shot of sight-pipe's view only represents a small part of the measured hole, it is necessary to take multiple shots in the circumferential direction and the depth direction. As shown in Fig. 3, the motion scheme can be described as Moving-Rotating-Moving plan. Begin at the bottom of the hole, rise to the top step-by-step and rotate a certain angle, then down to the bottom. Loop the procedure until covered the whole inner surface.

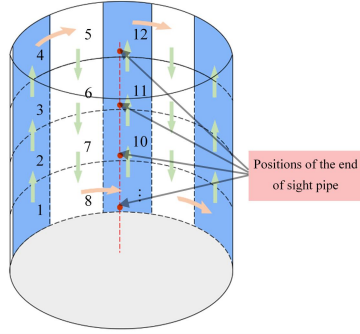

**Fig. 3** Moving-Rotating-Moving plan, the numbers denote the motion and capturing order

Assuming that the radius of the hole is $r$ and the depth of the hole is $h$, it can be calculated that the corresponding number of shots in the circumferential direction and depth direction by $[r/d_r]$ and $[h/d_h]+1$ respectively, where $[\cdot]$ denotes round down function. During image acquisition, the system drives the rotary stage and the translation stage to move corresponding times to capture the desired image.

## 3. ERROR ANALYSIS AND COMPENSATION

### 3.1. Projection error correction of arc surface

The inner surface of the hole is an arc, while imaging surface is a flat plane, which will introduce arc surface projection error. As shown in Fig. 4, the nearly parallel light irradiates the measured arc surface, and the resulting image is projected onto the image plane. Relationship between side $d_s$ of the expanded area and side $d_m$ of the imaging area is Eq. (3). The larger the radius of the hole, the smaller difference between $d_s$ and $d_m$, vice versa.

$$d_s = 2r\, arcsin(d_m/2r) \qquad (3)$$

Typically, let $r = 2.0\text{mm}$, $d_m = 2.53\text{mm}$, we get $d_s = 2.74\text{mm}$, and the corresponding error is 8.30%. To eliminate the projection error, we introduced a surface spreading method. After correction, such projection error could be ignored.

Let the center of image plane be original point $A$, the corresponding point of measured hole is $A'$. Starting from $A$, take any section of $AB$ along the horizontal direction, and its length is $w$, the corresponding arc on the measured surface is $\widehat{A'B'}$ with length $s$. The arc $\widehat{A'B'}$ of the measured hole is non-uniformly mapped to the image plane, and their relationship is $s = r\delta = r\, arcsin(w/r)$.

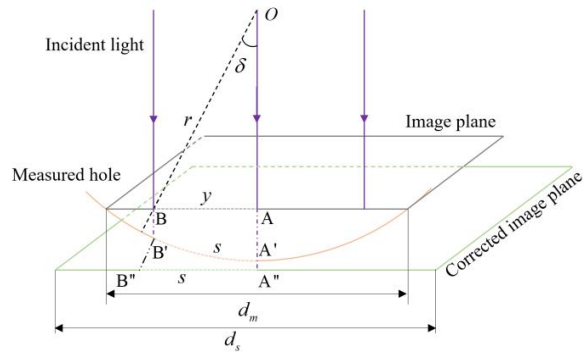

**Fig. 4** Projection error of the cylindrical inner surface to the image plane. Although the size of each pixel in the image of the imaging area is uniform, the size of the arc to be measured corresponding to each pixel is uneven.. According to the inverse mapping of $\widehat{A'B'}$ to $AB$, $AB$ is mapped to space where the pixel coordinates and the measured size are uniform, the corresponding area is $A''B''$.

An imaginary plane is introduced, which has the same length as arc length in the view field region, i.e. $\widehat{A'B'} = A''B''$. Note that w can be expressed as $w = kp_x$, where k is the number of pixels of B relative to the pixel of coordinate center A, and $p_x$ is the pixel equivalent (of physical distance) pixel/μm. We discrete the imaginary plane to the corrected image plane. In the corrected image plane, every pixel represents the same length of the measured surface. Hence the pixels near the center of image plane and corrected plane have the same pixel of equivalent of physical distance $p_x$, typically 2.16 μm/pixel. Assuming that the center of the corrected image plane is A'', and the pixel distance from A''B'' to A'' corresponding to AB is m. Since $\widehat{A'B'} = A''B''$, for a given w, we can get the corresponding m by Eq. (4). Therefore, the interpolated pixel k in image plane corresponding to pixel m in corrected image plane can be evaluated by Eq. (5). Besides, there is no influence of such projection error in vertical direction, just simply duplicate the pixel location of n.

$$m = s/p_x = r \cdot \arcsin(w/r)/p_x \tag{4}$$

$$k = r/p_x \cdot \sin(mp_x/r) \tag{5}$$

We employ bilinear interpolation to interpolate the corrected image plane. For any point (m, n) in corrected image, the corrected point can be evaluated by Eq. (6).

$$f(m,n) = f([m+1],[n+1])(m-[m])(n-[n]) + f([m+1],[n])(m-[m])([n+1]-n)$$
$$+ f([m],[n+1])([m+1]-m)(n-[n]) + f([m],[n])([m+1]-m)([n+1]-n) \tag{6}$$

The closer the point B to center point A, the smaller error between AB and $\widehat{A'B'}$, vice versa.

## 3.2. Manufacturing and assembly errors

Ideally, the axis of the sight-pipe should coincide with the axis of the hole. But in fact, it would deviate from the axis of the hole, due to the manufacturing error of mechanics and the installation. As shown in Fig. 5, the deviation can be divided into two cases, angle deviation and distance deviation.

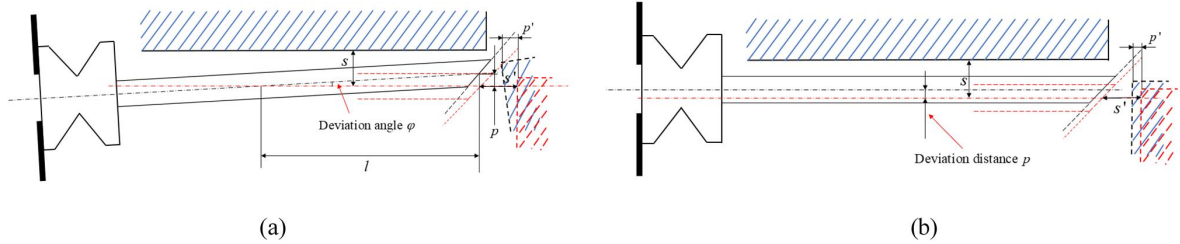

(a)          (b)

**Fig. 5** Two kinds of manufacture and installation error. (a) angle deviation. (b) distance deviation.

For angle deviation, assuming that the starting deviation point to the end of sight-pipe is l, the deviation of angle is φ. It would make the reflecting plane deviate $l \sin \varphi$. For distance deviation, the deviation of pipe shift is p. The total deviation of rotate and shift is $p_{total} = \sqrt{(l \sin \varphi)^2 + p^2}$. In our experiment, the upper bound of l is 45mm, the maximum deviation angle and distance deviation are not more than 0.5° and 0.2mm respectively. Hence, the total deviation $p_{total}$ is less than 0.44mm. Based on the microscope camera model, the minimum and maximum field of view can be defined as Eq. (7) and Eq. (8) respectively.

$$d_{min} = d_m - \frac{p_{total}}{r}(d_m - d_p) \tag{7}$$

$$d_{max} = d_m + \frac{p_{total}}{r}(d_m - d_p) \tag{8}$$

Define the maximum relative FOV range error $\varepsilon_m$ as Eq. (9).

$$\varepsilon_m = \max\{d_m - d_{min}, d_{max} - d_m\}/d_m = \frac{p_{total}}{r}(1 - d_p/d_m) \tag{9}$$

It can be calculated that for holes with a diameter of 4.0mm, $\varepsilon_m = 0.26\%$, for holes with diameter of 8.0mm, $\varepsilon_m = 0.22\%$. Therefore, the influence of manufacturing error of mechanics and installation error can be ignored.

### 3.3. Location of the defects

To better observe the surface of the inner hole, location and features are measured in every crack. For a certain hole, with inner diameter d and length l, it's easy to compute the angle of rotation $\alpha$ and moving step s under the condition of $fov_x = f_x$, $fov_y = f_y$, where $f_x$, $f_y$ represent the physical range of the effective region in the circumferential direction and depth direction. Then every effective region image can be located by $(j, k)$, where j represents the $j-th$ forward step and k represents the $k-th$ rotation. For every crack, its geometry center $(m, n)$ in pixel coordinate is computed, then it can be located by the map $(i, j, m, n, p_x, p_y) \rightarrow (z, \beta)$, where $p_x$, $p_y$ represent the x and y direction equivalent respectively, z represents the distance to the nozzle and $\beta$ represents the angle relative to the initial rotation. The area of cracks can be evaluated by $p_x$, $p_y$, $A_c$, where $A_c$ represents the pixel area of a crack in the image.

## 4. EXPERIMENT AND ANALYSIS

The measurement prototype is shown in Fig. 6, and the model of the system is described in detail in Fig. 1. The main components are listed below. LED light is composed of several LEDs, and the intensity of the lighting can be adjusted by controlling the current. The lens has a magnification between $\times 0.7 - \times 4.5$ and camera has high resolution. The translation stage motion straightness less than $20\mu m/100mm$, repeat positioning accuracy less than $3\mu m$. The rotary stage positioning accuracy less than 3', repeat positioning accuracy less than 18''. The surface roughness of the special sight-pipe holder is $20\mu m$. The parallelism between moving axis and rotating axis is guaranteed by the mechanic precision of cylindrical connector, cantilever and connecting plate. It is swift and efficient to measure fine holes with an inner diameter between 4mm and 6mm, a depth of no more than 47mm. The image acquisition and processing time does not exceed 10 minutes.

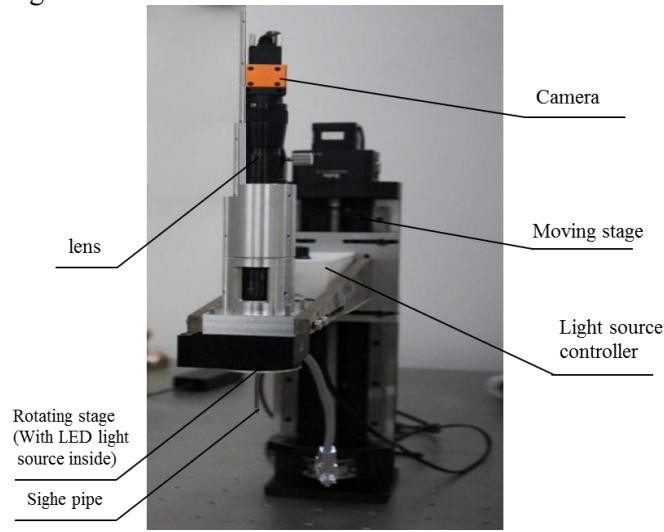

**Fig. 6** Prototype of the fine hole measurement system

As shown in Fig. 7, we make some cubes with the hole from the material to be tested, cutting it along the axis, and then punching some test holes and an engraved line in the groove as a test piece. The test hole and line are machined by high-precision machining tools. The diameter of the holes, width of the line and distance between holes are measured by optical coordinate measurement machine (CMM). We formulate a corresponding motion plan to drive the rotary stage, translation stage and sight-pipe to complete the image acquisition, then identify and locate the defects by our software on PC. Fig. 8 shows some of the detection consequence images of the simulated defects. Table 1 listed some of the results, Type I and II are tests of different benchmarks of diameter ø0.1mm and ø0.2mm respectively, Type III is the test of distance of 2 groups with 4 holes benchmark, Type IV is 0.3 mm line width benchmark test. Since The length of the reticle exceeds the field of view, the line width result is measured separately segment by segment and takes the averages. The result shows that in the benchmark test of ø0.1mm, ø0.2mm holes, 0.4mm hole distance and 0.3mm line width, the measurement standard deviation of the system is about $10\mu m$.

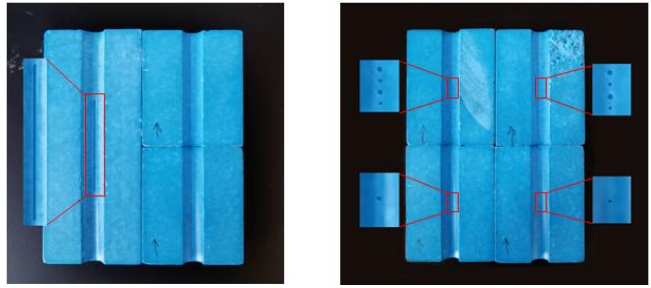

**Fig. 7.** The simulation of measured fine pipe and some of its cracks.

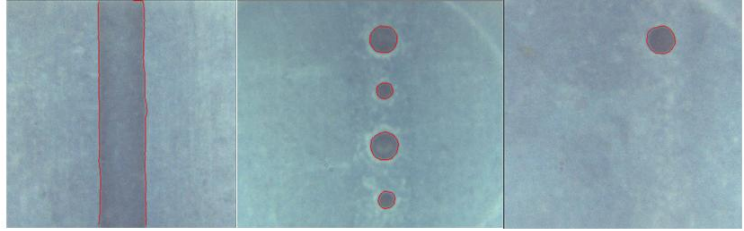

**Fig. 8** Detection results of simulating defects

Table 1 Measurement results of 4 kinds of benchmark compared with CMM

| Feature | CMM (Mean) (mm) | No. | Test value(Mean) (mm) | Standard deviation (mm) |
|---|---|---|---|---|
| Type I | 0.114 | #1 | 0.125 | 0.004 |
| | 0.122 | #2 | 0.115 | 0.002 |
| | 0.115 | #3 | 0.124 | 0.010 |
| | 0.114 | #4 | 0.123 | 0.011 |
| | 0.114 | #5 | 0.117 | 0.004 |
| | 0.114 | #6 | 0.120 | 0.007 |
| Type II | 0.205 | #1 | 0.202 | 0.004 |
| | 0.209 | #2 | 0.205 | 0.004 |
| | 0.207 | #3 | 0.202 | 0.006 |
| | 0.204 | #4 | 0.203 | 0.004 |
| | 0.202 | #5 | 0.203 | 0.003 |
| | 0.205 | #6 | 0.198 | 0.008 |
| Type III | 0.405 | #1 | 0.404 | 0.006 |
| | 0.399 | #2 | 0.396 | 0.004 |
| | 0.396 | #3 | 0.397 | 0.005 |
| | 0.394 | #4 | 0.394 | 0.005 |
| | 0.409 | #5 | 0.406 | 0.004 |
| | 0.387 | #6 | 0.385 | 0.006 |
| Type IV | 0.316 | #1 | 0.314 | 0.003 |
| | 0.314 | #2 | 0.312 | 0.005 |

## 5. CONCLUSION

In this paper, we designed an experimental prototype to fulfill the requirement of surface defect detection in fine holes. The sight-pipe is introduced to import external illuminating light into the hole and export the image of the measured surface at the same time. The arc surface projection error correcting method is proposed to make the system available for precision measurement. Experimental results show that the measurement standard deviations of the system for ø0.1mm, ø0.2mm hole diameter, 0.4mm hole spacing and 0.3mm line width are less than 10μm.The diameter of the tested hole can reach a range of 4mm to 6mm.The detection time for each hole is less than 10 minutes. This system is expected to be used for rapid detection of the inner surface of fine holes.